\documentclass[conference,compsoc]{IEEEtran}

\usepackage{multirow} 
\usepackage{url}
\usepackage{amssymb}
\usepackage{amsmath}
\usepackage{orcidlink}

\newcommand{\myparagraph}[1]{\vspace{1ex}\noindent{\bf #1}}
\usepackage{xspace}
\newcommand{\eg}{\textit{e.g.}\xspace}
\newcommand{\ie}{\textit{i.e.}\xspace}

\begin{document}
\title{Identifying and Mitigating Privacy Risks Stemming from Language Models}

\author{\IEEEauthorblockN{Victoria C. Smith\orcidlink{0000-0002-8629-0875}}
\IEEEauthorblockA{University College London, \\\& The Alan Turing Institute\\ London, UK \\ Email: v.smith.20@ucl.ac.uk}
\and
\IEEEauthorblockN{Ali Shahin Shamsabadi and Carolyn Ashurst}
\IEEEauthorblockA{The Alan Turing Institute\\ London, UK}
\and
\IEEEauthorblockN{Adrian Weller}
\IEEEauthorblockA{Unvisersity of Cambridge \\\& The Alan Turing Institute\\ London, UK}
}

\maketitle

\begin{abstract}
 Large Language Models (LLMs) have shown greatly enhanced performance in recent years, attributed to increased size and extensive training data. This advancement has led to widespread interest and adoption across industries and the public. However, training data memorization in Machine Learning models scales with model size, particularly concerning for LLMs. Memorized text sequences have the potential to be directly leaked from LLMs, posing a serious threat to data privacy. Various techniques have been developed to attack LLMs and extract their training data. As these models continue to grow, this issue becomes increasingly critical. To help researchers and policymakers understand the state of knowledge around privacy attacks and mitigations, including where more work is needed, we present the first SoK on data privacy for LLMs. We (i) identify a taxonomy of salient dimensions where attacks differ on LLMs, (ii) systematize existing attacks, using our taxonomy of dimensions to highlight key trends, (iii) survey existing mitigation strategies, highlighting their strengths and limitations, and (iv) identify key gaps, demonstrating open problems and areas for concern.
\end{abstract}

\section{Introduction}

There have been significant advances in recent years in developing and deploying Large Language Models (LLMs). This includes a trend towards increasingly large, data-hungry models pre-trained on extensive datasets. LLMs promise to bring real-world benefits to many fields, including those that rely on highly sensitive datasets, such as healthcare~\cite{yang2022large,benegas2022dna, li2021bioseq}. LLMs are now an integral part of many production systems, including dynamic chatbots, software writing, search engines, legal paraphrasing and drafting~\cite{shaghaghian2020customizing, chalkidis2020legal}, electronic health record~\cite{yang2022large} and DNA, RNA and protein analysis~\cite{benegas2022dna, li2021bioseq}. Public awareness and use of these models have also rapidly increased, evident in the widespread adoption of ChatGPT amongst the general public. 

However, LLMs present a range of privacy risks. It has been demonstrated that Machine Learning (ML) models can inadvertently memorize and disclose information from their training data, termed training data leakage~\cite{shokri2017membership, yeom2018privacy}, jeopardizing the privacy of the data they were trained on. In addition to privacy risks generally across ML models~\cite{song2021systematic, salem2023sok, rigaki2023survey}, LLMs bring unique privacy challenges~\cite{brown2022does} related to their training data, size, emergent abilities~\cite{wei2022emergent} and usage. 

Free text data can contain semantically similar private information in various formats, which brings unique challenges for effective data sanitization. Modern LLMs have millions to billions of parameters increasing their capacity for memorization~\cite{shokri2017membership, yeom2018privacy, carlini2019secret, carlini2021extracting, thomas2020investigating, ramaswamy2020training, carlini2022quantifying, mccoy2021much}. High numbers of parameters also bring specific challenges for efficiently implementing privacy-preserving measures (\eg Differentially Private Training). Emergent abilities are also attributed to vastly increased model parameters, resulting in qualitative changes in LLM behaviour. Emergent abilities unique to LLMs include: the ability to perform few-shot prompted tasks (or in-context learning)~\cite{brown2020language}, multistep reasoning~\cite{wei2022chain}, and self-calibration~\cite{kadavath2022language}, among others~\cite{wei2022emergent}.

It has also become increasingly common to train LLMs on large publicly available datasets and internet data, which poses privacy risks to individuals. For example, private information about an individual may be made public by someone else, or information may be shared for a narrow purpose without a reasonable expectation of it being used for model training~\cite{brown2020language}. Relatedly, LLMs can also make it easier to search and infer information about an individual from a wide range of online sources~\cite{staab2023beyond}. LLMs are commonly made available in full or via Application Programming Interfaces (APIs)  to large user bases who do not have direct access to the training data, raising concerns about training data leakage. Users of generative LLM chat models also can share sensitive data with the model~\cite{kshetri2023cybercrime}. This data is stored and may be used to update the LLM, and therefore has the potential for being leaked at a later date. 

As there is already a large and growing body of work regarding data privacy for LLMs, we present the first technical survey to clarify the progress made in evaluating and mitigating training data leakage. Our work categorises existing LLM privacy works across LLM architecture, size and development phase. We also discuss the emergent abilities of LLMs and comment on related privacy risks. Specifically, we make the following contributions:
\begin{itemize}
    \item We identify the salient dimensions where existing privacy attacks on LLMs differ. Namely, attacker goals, the development phase attacked, and the model architecture and size. 
    \item We survey existing attacks, and using our taxonomy of dimensions, we highlight key trends, current limitations and the most vulnerable development phases. 
    \item We discuss existing mitigation strategies, highlighting their strengths and limitations and identifying key gaps. 
    \item We conclude with a brief discussion, offering thoughts on open areas for research and areas for concern.
\end{itemize}

\section{Taxonomy of LLM Architectures \& Development Phases}

LLMs are concerned with learning the probability of tokens\footnote{Small meaningful units of text, which can be words, sub-words or characters, used as a functional semantic unit for processing text.} in a sequence of text based on their co-occurrences in the training corpora. LLMs are initially trained on large text corpora and are often referred to as ``foundation models"~\cite{bommasani2021opportunities}. Through transfer learning, LLMs have versatile applications in various downstream tasks, including \emph{Natural Language Understanding} (NLU) (\eg document classification, sentiment analysis, information extraction etc.)~\cite{devlin2018bert, wang2019extracting} and \emph{Generation} (NLG) (\eg question answering, text summarization, machine translation etc.). 

\subsection{Architectures}

LLMs primarily use transformer-based architectures~\cite{vaswani2017attention} which process the entire input sequence simultaneously and use an attention mechanism, which enables focus on particular tokens within a sequence and captures intra-dependencies across long sequences. Depending on the intended application, one of three types of transformer-based LLM architectures can be used:   

\myparagraph{(1) Encoder models} (also referred to as embedding models) use only the encoder component of the transformer model and are focused on generating contextual representations of input tokens. Input sequences are transformed into feature vectors, incorporating contextual information from the entire sequence (\ie bidirectional information). These feature representations find utility in various downstream NLU tasks~\cite{devlin2018bert, wang2019extracting}. Typically, a fully connected layer is added as the final layer of the model to facilitate these downstream tasks. Prominent examples include BERT~\cite{devlin2018bert}, DistilBERT~\cite{sanh2019distilbert} and RoBERTa~\cite{liu2019roberta}.

\myparagraph{(2) Decoder models} (often referred to as generative models) exclusively use the decoder component of the transformer. The decoder processes a sequence of tokens as input, generating a feature vector. The feature vector is used to calculate the probability distribution for the next token in the sequence. Each feature vector in this context is informed solely by the current token and the tokens that precede it. This unidirectional nature makes decoder models well-suited for NLG tasks. For this, the decoder takes input tokens and computes the probability distribution of the next token in the sequence. The output token is auto-regressive, reusing past outputs as inputs for subsequent steps. Noteworthy examples include GPT-2~\cite{radford2019language}, GPT-3~\cite{brown2020language}, GPT-Neo~\cite{black2022gpt}, LLaMa-2~\cite{touvron2023llama}, OTP~\cite{zhang2022opt}.

\myparagraph{(3) Encoder-decoder models.} (also generative models) initially, use an encoder to transform an input sequence into a series of continuous representations. These representations are then processed by the decoder, which generates an output sequence of tokens, one at a time, by calculating the most probable next token. This encoder-decoder framework is capable of handling sequence-to-sequence tasks with varying input and output sequence lengths, making it suitable for NLG tasks. Notable examples include T5~\cite{raffel2020exploring} and BART~\cite{lewis2019bart}.

\subsection{Development Phase}

The development of the Transformer architecture has ushered in a new era of very large LLMs with billions to trillions of parameters and a multi-phase training approach. 

\myparagraph{Pre-training.} Initially, transformers are typically pre-trained on a self-supervised language modelling task, which is used to learn the hidden weights of the model on unlabelled data. For example, pre-training encoder models involves randomly masking tokens in the input text sequence, with the model learning to recover these masked tokens in the output. Pre-training the decoder model involves computing the probability distribution of the next token. Pre-training typically utilizes publicly available, extensive and minimally-curated unlabeled text corpora, often scraped from the internet~\cite{raffel2020exploring, dodge2021documenting, brown2020language}. 

\myparagraph{Continued Pre-training.} Often, following pre-training on general domain data, LLMs are adapted to a specific domain by training the model on large domain-specific datasets (\eg PubMed) where available. This is often referred to as Continued Pre-training as the same self-supervised language modelling objectives are used and all model parameters are updated. Continued pre-training data is also often publicly available.

\myparagraph{Fine-Tuning.} LLMs are often adapted for specific downstream tasks by fine-tuning the model parameters on smaller task-specific datasets. This usually involves adding a task-specific layer to the pre-trained model and updating just the parameters in this layer, particularly when the fine-tuning dataset is small. Fine-tuning datasets are more likely to contain data not intended for public consumption. 

\myparagraph{In Context Learning}(ICL)~\cite{brown2020language} is an efficient alternative to fine-tuning which allows pre-trained generative LLMs to learn tasks from conditioning on a few examples given in a prompt, without updating model parameters. Typically a prompt is given to the model with example inputs and outputs to perform a task, alongside a test input. The model uses the information in the prompt to infer the latent concept shared by examples in the prompt and predict the next token~\cite{xie2021explanation, min2022rethinking}. Sensitive information regarding individuals or organisations has been entered in prompts to LLMs available via APIs. 

\section{Training Data Memorization in LLMs}

It has been demonstrated that ML models memorize aspects of their training data, and this memorized data can be extracted verbatim from the model itself~\cite{shokri2017membership, yeom2018privacy, carlini2019secret, carlini2021extracting, thomas2020investigating, ramaswamy2020training, carlini2022quantifying, mccoy2021much}, a phenomenon referred to as \emph{training data leakage}. Training data leakage can violate the privacy assumptions under which data was originally collected and make disparate information more easily searchable~\cite{staab2023beyond}. As LLMs have grown in size, the over-parameterization of models has made them particularly susceptible to such memorization~\cite{mccoy2021much, carlini2022quantifying}. 

Several works have investigated the memorization of training data in LLMs. Personal information (\eg URLs and phone numbers) or synthetically inserted ``secrets'' in training data can be extracted solely from the trained model~\cite{carlini2019secret, carlini2021extracting, thomas2020investigating, ramaswamy2020training}. For instance, GPT-2 was observed to memorize 600 out of 40GB of training data~\cite{carlini2019secret}, and it can reproduce passages from the training data exceeding 1,000 words~\cite{mccoy2021much}. More recently, experiments with larger GPT-J (6 billion parameters) revealed that it memorizes at least 1\% of the training data when prompted with training data prefixes~\cite{carlini2022quantifying}.

Memorization in LLMs has been found to scale with (1) model size (number of parameters)~\cite{carlini2022quantifying} and (2) duplicated text sequences in the training data~\cite{kandpal2022deduplicating, lee2021deduplicating, carlini2022quantifying}. The extent of memorized training data also increases with the number of tokens in the prompt given to an LLM~\cite{carlini2022quantifying}, meaning some memorization becomes noticeable only when the LLM is prompted with sufficiently long context. Additionally, recent examples encountered by LLMs during training are the most likely to be memorized~\cite{jagielski2022measuring}. This is particularly concerning for fine-tuning on private datasets.

\section{LLM Privacy Attack Threat Models}

Let $M$ be an LLM that outputs either an embedding or a probability distribution of the next token given an input text string or an input prefix. Let $D$ be a sensitive dataset. We investigate threat models of building $M$ using sensitive dataset $D$. 

\myparagraph{Parties.} The threat model usually considers two parties: a defender (i.e., an institution with the sensitive dataset $D$ and LLM $M$) and an attacker (i.e., an adversary whose goal is to leak information about $D$ through the services provided by the institutions based on $M$). The institution might use the sensitive dataset $D$ in various stages while building $M$, including pre-training, fine-tuning or compressing. 

\myparagraph{Privacy attacks.} Existing privacy attacks work in different threat models depending on the assumptions made on the knowledge and goal of the attacker. The knowledge of attackers about $M$ can be \emph{white-box} or \emph{black-box}. In the white-box threat model, the attacker has full knowledge of all components of $M$, including its architecture, parameters and gradients. The white-box setting is the best-case scenario for the attacker, as the attacker can design the attack by exploiting all components. While white-box access is less frequently encountered in practical situations, there are still conceivable scenarios where it could happen. For instance, this could occur if: (i) the attacker is an insider within an organization with access to the trained model, (ii) the entire model is made publicly available, and the distribution of the training data is known. More importantly, Federated Learning can be a real-world example in a white-box setting as each user and server observes model parameters. The knowledge of the attacker is limited to only the output of $M$ in the black-box setting. The black box is the worst-case scenario for the attacker as the attacker can only send arbitrary queries to $M$ and receive outputs, but they can not observe other information such as parameters and architecture of $M$. The black-box setting is the most common threat model in the real world. For example, consider services via public APIs. The attacker can access $M$ (pre-trained or fine-tuned) as an oracle, which takes an arbitrary sequence as input and outputs embeddings (in the case of masked language models) or prediction of the next token (in the case of autoregressive models). It also assumes the attacker can obtain or estimate probability vectors of the arbitrary sequences and, therefore, loss values. However, the adversary cannot obtain model weights or gradients. We can group the attacker's goal into two general categories of attacks: i) leaking information about data points in $D$ for example, inferring a sensitive attribute or absence/presence of a specific text data generated by a specific user; ii) reconstructing verbatim of a text sequence contributed by a user to $D$. 

\begin{table*}
\centering
\caption{Systematization of LLM privacy attacks in terms of attacker's access, type of the attack, number of shadow models, phase under attack, and LLM architecture. KEY-- WB: white-box; BB: black-box; MI: Membership Inference; AI: Attribute Inference; E: Extraction; PT: pretraining data; CPT: Continued pretraining data; FT: Fine-tuning data; ICL: ICL (prompt) data; \textsuperscript{$\dag$}: Although the study is defence-focused, there is an adversarial component. }
\small
\label{tab:llm_attacks}
\begin{tabular}{l|l|cc|ccc|c|cccc|ccc}
\hline

\multirow{2}{*}{\textbf{Year}}  &  \multirow{2}{*}{\textbf{Ref}} & \multicolumn{2}{c|}{\textbf{Access}} & \multicolumn{3}{c|}{\textbf{Attack}} & \textbf{\# Reference} & \multicolumn{4}{c|}{\textbf{Phase}} & \multicolumn{3}{c}{\textbf{Architecture}} \\ 

 &  & \textbf{BB} & \textbf{WB} & \textbf{MI} & \textbf{AI} &  \textbf{E} & \textbf{models}  & \textbf{PT} & \textbf{C-PT} & \textbf{FT} & \textbf{ICL} & \textbf{Enc.} & \textbf{Dec.} & \textbf{Enc-Dec} \\ \hline

2020 & ~\cite{hisamoto2020membership} & \checkmark & - & \checkmark & - & - & Multiple & \checkmark & - & - & - & - & - & \checkmark  \\ \hline

2020 & ~\cite{zanella2020analyzing} & \checkmark & - & - & - & \checkmark & 1 & - &  & - & - & \checkmark & - & -  \\ \hline

2021 & ~\cite{jagannatha2021membership}$\dag$ & \checkmark & \checkmark & \checkmark & - & - & 0 & - & \checkmark & - & - & \checkmark & \checkmark & - \\ \hline

2021 & ~\cite{carlini2021extracting} & \checkmark & - & - & - & \checkmark & 0 & \checkmark & - & - & - & - & \checkmark & -   \\ \hline

2021 & ~\cite{chen2021killing}$\dag$ & \checkmark & - & - & \checkmark &  - & 1 & \checkmark & - & - & - & \checkmark & - &  - \\ \hline

2022 & ~\cite{elmahdy2022privacy} & \checkmark & - & - & - & \checkmark & 0 & - & - & \checkmark & - & \checkmark & - & -  \\ \hline

2022 & ~\cite{zhang2023ethicist} & \checkmark & - & - & - & \checkmark & - & \checkmark & - & - & - & - & \checkmark & - \\ \hline

2022 & ~\cite{mireshghallah2022quantifying} & \checkmark & - & \checkmark & - & - & 1 & - & \checkmark & - & - & \checkmark & - & -  \\ \hline

2022 & ~\cite{carlini2022membership} & \checkmark & - & \checkmark & - & - & Multiple & \checkmark & - & - & - & - & \checkmark & - \\ \hline

2023 & ~\cite{wu2023privacy} & \checkmark & - & \checkmark & - & - & - & - & - & - & \checkmark & -& \checkmark & -  \\ \hline

2023 & ~\cite{tang2023assessing} $\dag$ & \checkmark & - & \checkmark & - & - & Multiple & - & - & \checkmark & - & - & - & \checkmark \\ \hline

2023 & ~\cite{li2023mope} & - & \checkmark & \checkmark & - & - & Multiple & \checkmark & - & - & - & - & \checkmark & -  \\ \hline

2023 & ~\cite{duan2023privacy}$\dag$ & \checkmark & - & \checkmark & - & - & 0 & - & - & \checkmark & \checkmark & - & \checkmark & - \\ \hline

2023 & ~\cite{fu2023practical}$\dag$ & \checkmark & - & \checkmark & - & - & 0 & - & - & \checkmark & - & - & \checkmark & -  \\ \hline

2023 & ~\cite{mattern2023membership} $\dag$ & \checkmark & - & \checkmark & - & - & 0 & - & - & \checkmark & - & - & \checkmark & -  \\ \hline

2023 & ~\cite{meeus2023did} & \checkmark & - & \checkmark & - & - & 0 & \checkmark & - & - & - & - & \checkmark & - \\ \hline

2023 & ~\cite{shi2023detecting}$\dag$ & \checkmark & - & \checkmark & - & - & 0 & \checkmark & - & - & - & - & \checkmark & -  \\ \hline

2023 & ~\cite{abascal2023tmi} $\dag$ & \checkmark & - & \checkmark & - & - & Multiple & - & - & \checkmark\footnote{Fine-tuned but infer data from pre-trained} & - & - & - & \checkmark  \\ \hline

2023 & ~\cite{staab2023beyond} $\dag$ & \checkmark & - & - & \checkmark & - & - & \checkmark & - & - & - & - & \checkmark & -  \\ \hline

2023 & ~\cite{al2023targeted} & \checkmark & - & - & - & \checkmark & 0 & \checkmark & - & - & - & - & \checkmark & -\\ \hline

2023 & ~\cite{morris2023language} & \checkmark & - & - & \checkmark & - & - & - & - & - & \checkmark & - & \checkmark & -  \\ \hline

2023 & ~\cite{nasr2023scalable} & \checkmark & - & - & - & \checkmark & - & \checkmark & - & - & - & - & \checkmark & - \\ \hline

2023 & ~\cite{yu2023bag} & \checkmark & - & - & - & \checkmark & 0 & \checkmark & - & - & - & - & \checkmark & - \\ \hline

2023 & ~\cite{al2023traces} & \checkmark & - & - & - & \checkmark & - & \checkmark & - & - & - & - & \checkmark & - \\ \hline

2024 & ~\cite{duan2024membership} $\dag$ & \checkmark & - & \checkmark & - & - & - & \checkmark & - & - & - & - & \checkmark & -  \\ \hline
\end{tabular}
\end{table*}

\begin{table*}
\centering
\caption{Systematization of LLM privacy attacks in terms of the type of the model, task, training dataset and attacker success.}
\footnotesize
\label{tab:AttackEval}
\begin{tabular}{l|l|p{20mm}|l|p{20mm}|p{30mm}} 
\hline

\textbf{Year}  &  \textbf{Ref} & \textbf{LLM Arch.} & \textbf{LLM Task}  &  \textbf{Data} & \textbf{Metrics} \\ 
\hline

2020 & ~\cite{hisamoto2020membership} &  6-layer Transformer & Machine Translation & WMT 2018 & Acc.  \\ \hline

2020 & ~\cite{zanella2020analyzing} &  BERT-like (4 layer) &  & Reddit comments & Differential Score  \\ \hline

2021 & ~\cite{jagannatha2021membership}$\dag$ &  BERT, DistilBERT, GPT2 & Language Modeling & MIMICIII, UMM, VHA & Membership Advantage (Empirical Privacy Leakage)   \\ \hline

2021 & ~\cite{carlini2021extracting} &  GPT-2 & Language Modeling & Common Crawl & Extracted Sequences Count  \\ \hline

2021 & ~\cite{chen2021killing}$\dag$ & BERT & Text Classification 	& TP-US, AG news, Blog dataset & Accuracy\\ \hline

2022 & ~\cite{elmahdy2022privacy} &  BERT-base & Text Classification & Reddit & Extracted Sequences Rate  \\ \hline

2022 & ~\cite{zhang2023ethicist} & GPT-Neo & Text Generation & LM-Extraction benchmark & Recall, Extracted Sequences Rate \\\hline

2022 & ~\cite{mireshghallah2022quantifying}  & ClinicalBERT, PubMedBERT, Bert-base-uncased & Language Modeling & MIMIC-III, i2b2 & AUC, Precision, Recall  \\ \hline

2022 & ~\cite{carlini2022membership} & GPT-2 & Text Generation & WikiText-103 & TPR, FPR   \\ \hline

2023 & ~\cite{wu2023privacy} &  GPT-3 & Text Classification & SST-2, Amazon, AG-news, TREC & Accuracy \\ \hline

2023 & ~\cite{tang2023assessing}&  BART-base, FLAN-T5 base & Text Summarization & SAMsum,  CNN/DailyMail, MIMIC-cxr & Accuracy, AUC,TPR@low\%FPR   \\ \hline

2023 & ~\cite{li2023mope} & Pythia & Text Generation & The Pile & AUC, TPR@low\%FPR   \\ \hline

2023 & ~\cite{duan2023privacy} &  GPT-2 & Text Generation & GLUE & AUC, TPR@low\%FPR   \\ \hline

2023 & ~\cite{fu2023practical} & GPT-2, GPT-J, Flacon-7B, Llama-7B & Text Generation & Wikitext-103, AG News, XSum, Wikicorpus, TLDR News, CNNDM & AUC   \\ \hline


2023 & ~\cite{mattern2023membership}  &  GPT-2 & Text Generation & AG News, Sentiment140, Wikitext-103 & TPR\@low\%FPR   \\ \hline

2023 & ~\cite{meeus2023did} & OpenLLaMA-7B, OpenLLaMA & Text Generation & Arxiv, RedPajama & AUC, TPR@low\%FPR   \\ \hline

2023 & ~\cite{shi2023detecting} & LLaMA, GPT-NeoX, Pythia & Text Generation & WIKIMIA & AUC, TPR@low\%FPR   \\ \hline


2023 & ~\cite{abascal2023tmi}  &  Transformer-XL & Text Classification & WikiText-103 & TPR@low\%FPR   \\ \hline

2023 & ~\cite{staab2023beyond}  &  Llama-2 (7B, 13B, 70B), GPT-(3.5, 4), PaLM 2-(text, Chat), Claude-(2, Instant) & - & PersonalReddit & Accuracy   \\ \hline

2023 & ~\cite{al2023targeted} &  GPT-Neo & Text Generation & SATML2023 & Precision, Recall at 10\% FPR   \\ \hline

2023 & ~\cite{morris2023language} & Llama-2-(7b, Chat) & - & Instructions-2M & Token F1-score, BLEU score, Exact Match   \\ \hline

2023 & ~\cite{nasr2023scalable} &  GPT-Neo, Pythia, RedPajama-INCITE & Text Generation & The Pile, Red Pyjama & Extracted Sequences Rate   \\ \hline

2023 & ~\cite{yu2023bag} &  GPT-Neo 1.3B & Text Generation & The Pile & Precision, Recall, Hamming Distance   \\ \hline

2023 & ~\cite{al2023traces} & GPT-NEO, GPT-2, Pythia, CodeGen & Text Generation & Code Dataset & Exact Match, BLEU score   \\ \hline

2024 & ~\cite{duan2024membership}  & PYTHIA, GPT-NEO & Text Generation & The Pile & AUC, TPR@low\%FPR   \\ \hline
\end{tabular}
\end{table*}

\section{LLM Privacy Attack Systemizations.} Table~\ref{tab:llm_attacks} systematically classify existing LLM privacy attacks based on threat models, attack types and our LLM taxonomy in terms of the LLM architecture and phase in which private dataset has been used. Table~\ref{tab:llm_attacks} covers 22 LLM privacy attack-focused studies and divides them into three main types of privacy attacks: i) \textbf{Membership inference Attacks} (MIAs) aim to infer whether a particular data point was used to train a model; ii) \textbf{Attribute Inference attacks} aim to deduce sensitive aspects of the training data; and iii) \textbf{Data Extraction Attacks} aim to reconstruct verbatim text sequences from the training data. 
Table~\ref{tab:AttackEval} analyses dataset, models, tasks and metrics considered in each of these LLM privacy attacks.

\section{Membership Inference Attacks on LLMs}
\label{subsection: MIAs}

Membership Inference Attacks (MIAs) aim to infer whether a particular text instance was part of a target model's training dataset~\cite{shokri2017membership}. Such attacks can lead to various privacy violations~\cite{tabassi2019taxonomy}.  For example, \textbf{identifying that a text sequence generated by a Clinical LLM (trained on Electronic Health Records) is likely part of the training data can reveal information about a patient which violates their patient rights}~\cite{jagannatha2021membership}.

\looseness=-1 Formally, the goal of the MIA is to infer if a given text instance $x$ was part of the training dataset $D$ for the LLM $M$ through computing a membership score $f(x; M)$. A specific threshold, $t$, is utilized on the membership score to ascertain membership status. \textbf{We group existing MIAs based on their scoring functions into eight categories}, namely Reference-free Loss-based MIAs, Reference-based Loss-based MIAs,  Likelihood Ratio Attacks MIAs, Zlib Entropy MIAs, Min-$k\%$ Prob MIA, Data Perturbation-based MIAs, Model Perturbation-based MIAs and Paraphrasing-based MIAs.  

\myparagraph{(1) Reference-free Loss-based MIAs}~\cite{yeom2018privacy, jagannatha2021membership} exploit models’ tendency to overfit their training data and, therefore, exhibit lower loss values for training members. Loss-based MIAs define their scoring function, $f_{\text{Loss-based}}$, based on the loss, $L(x; M)$, of the target text instance under the target model as $f_{\text{Loss-based}}(x; M) = L(x; M)$. Loss-based MIAs predict membership as follows: if {$f_{\text{Loss-based}}(x; M) \le t$}, where $t$ is the specified threshold, the adversary rejects the null hypothesis, assuming $x$ is a member of the training data of $M$. Loss-based MIAs stage on LLMs achieve low success~\cite{jagannatha2021membership, carlini2022membership}. This can be attributed to certain samples (\eg simple short sentences or repetitive samples) being assigned higher probabilities than others, and the influence of this aspect on the obtained model score largely outweighs the influence of a model’s tendency to overfit its training samples. In summary, most reference-free MIAs yield a low success, which is only slightly better than random guesses.
    
\myparagraph{(2) Reference-based Loss-based MIAs} address the above limitation of reference-free loss-based MIAs by calibrating model outputs using reference models~\cite{carlini2021extracting}, $M_{\text{ref}}$. Reference models are trained on training data drawn from a similar distribution to the training dataset $D$ of the model $M$ under the attack. The logic behind this method is that $M_{\text{ref}}$ can provide perspective on the degree to which a particular sample deviates from the probability distribution of the target model~\cite{long2018understanding, watson2021importance}. The reference-based Loss-based MIA considers the membership score of the target text instance $x$ by $M$ relative to the membership score of the target text instance $x$ by of $M_{\text{ref}}$. Reference-based Loss-based MIAs define their membership scoring function, $f_{\text{Ref\&Loss-based}}(x;M) = L(x;M) - L(x; M_{\text{ref}})$. The Reference-based Loss-based MIAs are more effective than Reference-free Loss-based MIAs in LLMs as Reference-based Loss-based MIAs measure a more reliable membership signal by comparing the probability discrepancy between the target model and the reference model. Both Reference-free Loss-based MIAs and Reference-based Loss-based MIAs are typically staged under a black-box threat model~\cite{carlini2021extracting, mireshghallah2022quantifying, carlini2022membership, mattern2023membership}. However, reference-based loss-based MIAs assume access to data from the same underlying distribution as the training data to create reference models. 

\looseness=-1 \myparagraph{(3) Likelihood Ratio Attacks MIAs} (LiRA)~\cite{carlini2022membership} treat the MIA as a hypothesis test using the likelihood ratio between the loss distribution, $Q_{in}$, over models trained on point $(x, y)$,  compared with those not trained on $(x,y)$, $Q_{out}$: $f_{\text{LiRA-based}}(x;M) = \tilde{Q}_{in/out}(x, y)$. They estimate the distribution of losses for each distribution using shadow models either trained on $(x, y)$ or not. This only requires black-box access but assumes that an adversary has knowledge about the distribution of the target model’s training data and the strong assumption of access to a sufficient amount of samples from it. LiRA has also been adapted to Encoder Masked LMs~\cite{mireshghallah2022quantifying}, which do not explicitly define an easy-to-compute probability distribution over natural language sequences, as follows: (1) estimating the likelihood ratio for an input sequence by masking 15\% of tokens in the sequence at random, $k$ times to generate a set of masking patterns for the sequence (2) passing each combination of masking patterns through the model to calculate the output scores (3) combining outputs to give the probability distribution over the sequence. 

\myparagraph{(4) Zlib Entropy MIAs}~\cite{carlini2021extracting} relaxes the data-related assumption, which is needed in Reference-based Loss-based MIAs. Instead, Zlib Entropy MIAs calibrate the loss $L(x; M)$ by calculating the zlib entropy of the text instance $x$, which is the number of bits of entropy when a text sequence $x$ is compressed with zlib compression~\cite{gailly2004zlib}. Zlib Entropy MIAs define their scoring function, $f_{\text{zlip}}(x;M) = \frac{L(x;M)}{\text{zlib}(x)}$, as the ratio between the loss and the zlib entropy of the text instance, 
where $\text{zlib}$ compresses and filters out common text sequences. 

\myparagraph{(5) Min-$k\%$ Prob MIA}~\cite{shi2023detecting} is another reference-free method that calibrates $L(x; M)$ based on the fact that i) a non-member text instance most likely contains a few outlier tokens with low likelihoods under the LLM; ii) a member text instance is less likely to have words with such low likelihoods. Therefore, Min-$k\%$ Prob MIA uses the $k\%$ of tokens with the lowest likelihoods, forming a set $\text{Min-}k\%(x)$, to compute a score for the sequence $x$ instead of averaging over all token probabilities as in loss, $f_{\text{Min-k\% Prob}}(x;M) = \frac{1}{|\text{Min-}k\%(x)|}\sum_{x_{i}\in \text{Min-}k\%(x)} - \log(p(x_{i}|x_{1},...,x_{i-1}))$.
Min-$k\%$ Prob MIA does not require knowledge of the training data, works in black-box threat models, and does not need training any reference model. 

\myparagraph{(6) Data Perturbation-based MIAs}~\cite{mattern2023membership, fu2023practical,li2023mope} calibrate the loss by estimating the curvature of the loss function at the given text instance. The logic behind this attack is that in a well-generalised model, a training point may not have a lower absolute loss level than a test point. To do this, model outputs for a given sample are compared to outputs of perturbed neighbour texts $\{\tilde{x}_i\}_{i=1}^n$~\cite{mattern2023membership}. The neighbouring samples are textual samples crafted to be non-training members that are as similar as possible to the target point. The average probability of neighbour sequences under the target model can then be used to calibrate the probability of the target sequences as, $f_{\text{Perturbation-based}}(x, M\theta) = L(x; M) -\frac{1}{n}\sum^{n}_{i=1}L(\tilde{x}_{i};M)$. This can also be combined with a reference model~\cite {fu2023practical} to calibrate the probabilistic variation measured on the target model. 

\looseness=-1 \myparagraph{(7) Model Perturbation-based MIAs}~\cite{li2023mope} are an alternative to Data Perturbation-based MIAs in a white-box setting. Model Perturbation-based MIAs perturb the LLM parameters with Gaussian noise to estimate the loss landscape at a given point $x$. 

\myparagraph{(8) Paraphrasing-based MIAs}~\cite{fu2023practical} use the probability variation outputted by $M$ to infer membership. Paraphrasing-based MIAs are designed based on the intuition that memorized text instances typically live in the domains of local maxima of the probability function parameterized by an LLM. Paraphrasing-based MIAs use a paraphrasing model and generate paraphrased text close to the target text instance $x$ in the probability distribution for calculating probabilistic variations. This type of MIA does not rely on overfitting as compared to using the loss-based MIAs. 

\vspace{-3mm}
\subsection{Analysis of Membership Inference Attacks on LLMs}
We lay out questions to understand the factors impacting the effectiveness of MIAs, and answer them based on the existing works.

\myparagraph{Perturbation-based MIAs versus Loss-based MIAs or their combination?} Model Perturbation-based MIAs which perturb model parameters~\cite{li2023mope}, are inherently white-box. Perturbation-based MIAs outperform loss-based MIAs when the attacker has white-box access to the Pythia Suite of LLMs~\cite{li2023mope}. The combination of perturbation-based and loss-based MIAs can result in a stronger MIA~\cite{li2023mope}, especially for larger model sizes. This raises the idea that different test statistics could be combined in specific scenarios to yield stronger attacks. Although white-box perturbation-based MIAs do not require the adversary to train a reference model, the white-box assumption is not always a realistic assumption, with API access to LLMs being common. Data Perturbation-based MIAs, which perturb input instead of model parameters, can work in black-box settings, which is a more realistic scenario. Also, Data-perturbation-based MIAs do not require access to data from the training data distribution. In practice, to create neighbours, they perturb the text instance for example using BERT to replace a percentage of randomly selected token spans while maximizing the neighbour's likelihood~\cite{mattern2023membership}. Perturbation-based MIAs can be combined with a reference LLM trained on data from the same domain as the training data to improve further the attack success~\cite{fu2023practical}. This data can be constructed by prompting the target generative LLM with short text chunks and collecting the associated outputs. 

\looseness=-1 \myparagraph{How does LLM size affect vulnerability to MIAs?}
LLM size has a different impact on the vulnerability of LLMs to MIAs depending on the type of attack. The performance of most types of MIAs, including loss-based and min-$k\%$ MIAs, increases as we increase the size of LLMs~\cite{shi2023detecting,jagannatha2021membership,duan2024membership}. For example, the AUC score of min-$k\%$ reference-free MIA on detecting pretraining 128-length texts from different-sized LLaMA models (7-65B) rises with increasing LLaMA size~\cite{shi2023detecting}. This is because larger LLMs have more parameters and thus are more likely to overfit and memorize the training data. However, model perturbation-based MIAs show an opposite trend~\cite{li2023mope}.

\myparagraph{How does LLM architecture affect vulnerability to MIAs?} \cite{jagannatha2021membership} evaluated vulnerability of BERT, DistilBERT and GPT2 to MIAs in both white-box and black-box settings. The black-box MIAs are based on a loss-based MIA~\cite{yeom2018privacy}. However, the white-box MIA exploits aggregated gradient values from each model layer by taking the squared norm of all parameters in each layer. GPT2 consistently has the highest leakage, and DistilBERT has the lowest, likely due to the lower number of parameters in the model and, therefore, lower capacity for memorization~\cite{carlini2022quantifying}. 

\myparagraph{How does fine-tuning affect the success of MIAs on pre-training data?} We consider two scenarios: 1) the attacker has black-box access to both the pre-trained model and the fine-tuned model; 2) the attacker has black-box access only to the fine-tuned model. In both of these scenarios, the attacker's goal is to infer information regarding the membership of data points in the training set of the pre-trained model. In scenario (1), the attacker can use the difference in probabilities of the sequence under the pre-trained model and fine-tuned models (or differential score) as the membership score~\cite{zanella2020analyzing}. 
TMI~\cite{abascal2023tmi} investigates scenario (2) where the adversary has access to only a fine-tuned model and tries to make inferences back to the pre-trained model. In this scenario, a Transformer-XL model is pre-trained on WikiText-103 but fine-tuned on the DBpedia ontology classification dataset. The attacker's goal in TMI is to infer information regarding WikiText-103 by only having black-box access to the fine-tuned Transformer-XL. TMI works by constructing a dataset of prediction vectors from queries to shadow models of the fine-tuned model to train a metaclassifier that can infer membership. TMI achieved an AUC score of 0.78 under black-box access to the fine-tuned model. 

\myparagraph{Does fine-tuning different parts of an LLM have the same impact on the success of MIAs on the fine-tuning data?} An LLM is usually fine-tuned on a private dataset in three different ways: (1) fine-tuning all layers, (2) fine-tuning a few layers, especially the last layer of the model, and (3) fine-tuning adapters (i.e., small bottleneck modules inserted within transformer blocks). (1) and (2) are because of the fact that fine-tuning all the model parameters can be computed and memory-intensive~\cite{fedus2022switch}. ~\cite{mireshghallah2022memorization} analysed the privacy risks of wikitext2-raw-v1, Penn Treebank or Enron Email used in fine-tuning GPT-2 based on each of the above approaches through black-box and reference-based MIAs. Fine-tuning the last layer of the model has the highest susceptibility to MIAs~\cite{mireshghallah2022memorization}. This observation implies a large number of parameters, and the location of parameters at the end of the model creates a high capacity to memorize the fine-tuning data. 

\myparagraph{How to evaluate the success of MIA when the training data of the victim model is unknown?} In some practical scenarios, details of the training data of the victim model have not been released at all (\eg GPT-3) making it challenging to prepare a dataset of members and non-members for evaluating the success of MIAs. The only side knowledge that the attacker could have is the year when the victim models were trained. We can create the evaluation dataset by curating data from Wikipedia and considering those published before the year that the model was trained (e.g., 2023) as the members of training data and after 2023 as the non-members~\cite{shi2023detecting}. 

\myparagraph{How to improve the success of reference-based MIAs?} Reference-based MIAs are the most successful attacks across pre-trained LLMs from the PYTHIA model suite and the GPT-NEO family, trained on the Pile, knowledge sources (Wikipedia), academic papers (PubMed Central, ArXiv), dialogues (HackerNews), and specialized-domain datasets (DM Math, Github)~\cite{duan2024membership}. However, the AUC scores of Reference-based MIAs do not exceed 0.6 for any domain. The success of Reference-based MIAs can be improved by enhancing the quality of the reference models. The quality of the reference models can be enhanced in two main ways; namely, i) using the same architecture as the LLM under the attack~\cite{tang2023assessing}, and iii) using a reference dataset that is more similar to the target dataset~\cite{fu2023practical}. However, in practice, it is not always possible to satisfy these requirements due to the architecture and training dataset of the target model being hidden, resulting in difficulties in selecting reference models.  

\looseness=-1 \myparagraph{At which level of text can we infer membership?} MIAs are staged to infer membership of a sequence~\cite{tang2023assessing} and a document~\cite{meeus2023did} in different tasks such as summarization. ~\cite{tang2023assessing} explore reference model MIAs for data summarization tasks in sequence-to-sequence models (BART and FLAN-T5). The adversary trains a reference model on data with a similar underlying distribution and uses the known data membership of the shadow models as training labels to train an attack classifier, whose goal is to predict the data membership of the reference model. 

~\cite{meeus2023did} extend MIAs to document-level in a black-box threat model on pre-trained OpenLLaMA-7B using books and academic papers from the RedPajama dataset. Their approach involves querying the LLM to retrieve the predicted probability of a token given a context of a specified length. They carry this out for different context windows to obtain multiple probability values and then normalise these values based on the rarity of the token. They then aggregate all the token-level information for a document to construct a document-level feature. They use two aggregation methods for this: AggFe and HistFE. This document-level feature is passed as input to a binary classification model to predict membership. They achieved a strong AUC of 0.856 for books and 0.678 for papers on OpenLLaMA-7B, outperforming sentence-level documents adapted for the same task. They find the smaller OpenLLaMA-3B similarly vulnerable to their approach. 

\myparagraph{Which one is safer for adapting LLMs to a private domain: in-context learning or fine-tuning?} In-context learning~\cite{brown2020language} is an efficient alternative to fine-tuning in which an LLM is augmented with private data. In-context learning does not update model parameters. However, it introduces privacy risks for those data used for the augmentation~\cite{tang2023privacy,duan2024flocks,wu2023privacy}. In-context learned LLMs are more vulnerable to MIAs than fine-tuned LLMs~\cite{duan2023privacy}. This observation is only for a controlled and limited environment. 

\myparagraph{Which one is less inferable: pre-training or fine-tuning data?} Overall, the threat of MIAs is low in pre-trained LLMs due to two typical characteristics of pre-trained LLMs: (1) large training data size (typically LLMs are trained with billions to trillions of tokens) and (2) low number of training epochs (typically LLMs are trained for ~one epoch due to the large dataset size and their overfitting tendency). However, fine-tuning data, which tends to be smaller and in training for more epochs, has been shown to be highly vulnerable, particularly if just the last layer of the model is fine-tuned~\cite{mireshghallah2022quantifying}. 

\myparagraph{How do characteristics of data(set) affect its susceptibility to MIAs?} Properties such as length of text instances, dataset size~\cite{jagielski2022measuring}, data training order~\cite{li2023mope,biderman2023emergent} and data duplication~\cite{kandpal2022deduplicating} affect the susceptibility of text instances to MIAs.
The AUC score of MIAs increases as text length increases~\cite{shi2023detecting,mireshghallah2022quantifying}. This is likely because longer texts contain more unique information memorized by the target model, making them more distinguishable from unseen texts. As the size of the training set increases, the success of MIAs decreases monotonically~\cite{tang2023assessing,shi2023detecting}. This suggests that increasing the number of samples in the training set can help to alleviate overfitting and reduce the MIAs AUC score. \cite{duan2024membership} empirically analyse reference-based MIAs across LLMs with different training data, showing MIA performance spikes before gradually decreasing as the amount of training data seen increases. They speculate that the spike and then a gradual decline in performance is because the data-to-parameter-count ratio is smaller early in training, and the model may tend to overfit but generalize better as training progresses, reducing attack capability. Regarding other properties, such as data training orders and data duplication, it is still not clear how they impact the performance of MIAs. There have been several contradictive observations in the literature regarding this impact. For example, ~\cite{jagielski2022measuring} shows that, particularly on large data sets, the success of loss-based MIAs increases for text instances used in more recent batches. However, ~\cite{li2023mope} and ~\cite{biderman2023emergent} found no correlation between the order of text instances in training and their susceptibility to MIAs. Additionally, the success of MIAs against LLMs has been demonstrated to be correlated with the presence of duplicated sequences in the training data~\cite{kandpal2022deduplicating,duan2024membership}. However, ~\cite{kandpal2022deduplicating} found that the reference-based MIAs perform almost equally as well on both models trained on non-deduplicated datasets and deduplicated datasets.

\section{Data Extraction Attacks on LLMs}
\label{subsection: data extraction}

Data extraction attacks~ aim to extract memorized training text instances from a language model~\cite{carlini2021extracting,yu2023bag,zhang2023ethicist,nasr2023scalable,al2023traces,elmahdy2022privacy,kandpal2022deduplicating,carlini2022quantifying,mireshghallah2022memorization}.  Such \textbf{data extraction attacks on LLMs can lead to various privacy violations as training data may contain sensitive information such as names, email addresses, phone numbers, and physical addresses.}

\myparagraph{Categorization of data extraction attacks on LLMs.} Data Extraction Attacks can be \textit{untargeted}~\cite{carlini2021extracting} or \textit{targeted}~\cite{al2023targeted,zhang2023ethicist} with respect to text instances that the adversary is searching to recover. Untargeted data extraction attacks aim to recover any memorised text instances. Therefore, untargeted data extraction attacks might extract uninteresting text instances. However, targeted data extraction attacks aim to recover the suffix for a specific prefix that the adversary has access to. Therefore, targeted data extraction attacks can recover specific information from the training data of the LLM. For example, an attacker can perform targeted reconstruction attacks to complete a private email that they know partially i.e., the beginning of the email~\cite{zhang2023ethicist} or the phone number of a specific person by using the following prefix ``Yu's phone number is''~\cite{yu2023bag}.

\myparagraph{How untargeted data extraction attacks on LLMs work?} Given a LLM $M$ trained on private training data $D$, untargeted data extraction attacks contain three steps~\cite{carlini2021extracting}: i) \textbf{Prefix generation.} Prefixes can be initialized using data from Common Crawl to ensure the absence of unusual prefixes~\cite{carlini2021extracting} or they can be initiated as empty; ii) \textbf{Text Candidate Generation.} A large amount of text instances are generated by unconditionally sampling from $M$, when the model is conditioned on some prefix consisting of a sequence of tokens $\{x_1, ..., x_i\}$.  Deploying a decaying temperature setup can increase diversity in early token predictions and increase predictability as the generation progresses~\cite {carlini2021extracting}; and iii) \textbf{Text Canidate Selection.} Generated suffixes are deduplicated and sorted in order of likelihood to retain only the most likely memorized text instances using membership inference attacks (MIAs) (detailed in Section~\ref{subsection: MIAs}). Using strong MIAs such as reference models and zlib entropy can improve the candidate selection. 

\myparagraph{How targeted data extraction attacks on LLMs work?} Given $M$ trained on $D$ and a prefix, targeted data extraction attacks contain two steps~\cite{zhang2023ethicist}: i) Suffix Candidate Generation for the Given Prefix; and ii) Suffix Candidate Selection. 

The generation's likelihood of the ground truth suffix can be increased using loss-smoothed prompt-tuning to train the soft prompt tokens and reduce the difficulty of sampling the ground truth suffixes~\cite{zhang2023ethicist}. Sorting and normalising candidate suffixes relative to other candidates for a given prefix (as opposed to likelihood alone) can give a better signal for selecting the ground truth suffix~\cite{zhang2023ethicist}. Using these two improvements, ~\cite{zhang2023ethicist} extract 50-token suffixes from 50-token prefixes, achieving a recall score of 62.8\%, compared to baselines of using perplexity alone or calibrating with zlib or a reference model (all around 50\%). \cite{al2023targeted} use GPT-Neo and generate ten suffixes for a given prefix. They filter for samples with only the lowest loss per prefix. This allows them to extract the suffix for 69\% of the samples. In the second step, they use a classifier-based MIA. Overall the two-step approach reached 0.405 recall at 10\%FPR, improving on the baseline by 34\%. 

\myparagraph{Are data extraction attacks on LLMs practical?}
Recently, there have been some efforts demonstrating that data extraction attacks on LLMs are practical through attacking pre-trained GPT-2~\cite{carlini2021extracting}, GPT-Neo trained on the Pile~\cite{yu2023bag,al2023targeted}, closed-source (ChatGPT (gpt-3.5-turbo-instruct)), semi-open (LLaMa and Falcon), and open-sourced (Pythia and GPT-Neo) pre-trained autoregressive LLMs~\cite{nasr2023scalable}, MLM, BERTbase, fine-tuned for 10 epochs for text classification on the Reddit dataset~\cite{elmahdy2022privacy}. ~\cite{al2023traces} performed data extraction attacks on code generation models CodeParrot, CodeGen-Mono-16B and PyCodeGPT. ~\cite{nasr2023scalable} staged an extraction attack on closed-source (ChatGPT (gpt-3.5-turbo-instruct)), semi-open (LLaMa and Falcon), and open-sourced (Pythia and GPT-Neo) pre-trained autoregressive LLMs.  To do this, they started by downloading Wikipedia data and randomly sampling 5-token snippets. They then prompted the LLM with these prefixes to generate a 50-token string. They repeated this process 109 times per LLM. They measure memorization against the open-source dataset or, in the case of closed-source models, an approximate dataset they construct. They find ChatGPT is not very susceptible in this case. They find models tuned to be conversational are less likely to emit memorized sequences, which they postulate is down to model alignment tuning. Model alignment through instruction-tuning or reinforcement learning from human feedback can be used on production LMs to improve the conversational responses of LLMs. The adaptation to dialogue does not give the user direct control over the language modelling task. They propose a prompting strategy to make the LLM behave like a base language model and output more internet-style text by, for example, asking it to repeat the word poem forever. The model eventually diverges, and from these generations from ChatGPT, they found a small proportion of its generations are directly copied from the pertaining data. They found memorization was higher than expected with ChatGPT, one of the worst, with 0.852\% of generated tokens being part of 50-token sequences from the pretraining data. They also find model families that are trained for longer (\eg LLaMA) memorize more than model families trained for less long (\eg OPT), relative to their size. 

~\cite{al2023traces} performed data extraction attacks on code generation models CodeParrot, CodeGen-Mono-16B and PyCodeGPT. They found these are also vulnerable to extraction of up to 47\% of code used in training. The noted code-trained LLMs emit training data at a lower rate than natural language-trained LLMs. In line with other studies, they find the rate of memorisation scales with the model size and that duplication of training data increases the rate of memorisation~\cite{kandpal2022deduplicating, carlini2022quantifying}. They did also find examples of personal information being memorized by code models (\eg names, emails and usernames). Interstingly, they find each model family tested memorizes different sets of examples. 

\myparagraph{How does the training data affect the attacks?} ~\cite{kandpal2022deduplicating} carried out data extraction attacks in commonly used web-scraped training sets, demonstrating that in language models, the efficacy of data extraction can be frequently attributed to duplicate sequences in the training data. ~\cite{zhang2023ethicist} find that attack performance grows exponentially with prefix length across all methods but show that their method is consistently best for all prefix lengths. Likewise, the attack performance decreases as the predicted suffix length increases. \cite{inan2021training}, in a study on memorization, highly repeated sequences in the training data are the most likely to be leaked in extraction attacks. ~\cite{carlini2022quantifying} report the proportion of extractable sequences increases log-linearly with the number of prefix tokens. 

\myparagraph{Can training data extraction be adapted for Encoder LLMs?} \cite{elmahdy2022privacy} adapt the extraction attack to an encoder model, BERTbase, fine-tuned for 10 epochs for text classification on the Reddit dataset. Given a partial sequence with missing tokens, the extraction algorithm aims to enumerate all tokens from the vocabulary such that the corresponding class label achieves the highest likelihood under the target model. They constructed out-of-distribution canary sequences by randomly sampling tokens and inserting these into the dataset with several repetitions. The algorithm can only reconstruct two tokens successfully as the search space becomes too large after that. They found that high repetition of canaries improves success rate, in line with studies showing memorization is exacerbated by duplication~\cite{kandpal2022deduplicating, carlini2022quantifying}. Repetition still resurfaces the missing token if the algorithm generates more candidates (i.e., larger beam size). Inserting canaries into the rarest class reduces the effectiveness of extraction attacks, but placing similar canaries with some small token perturbations into more prevalent classes makes attacks more effective.  

\section{Attribute Inference Attacks on LLMs} 

Attribute inference attacks aim to deduce sensitive attributes about an individual from a partially known training record~\cite{fredrikson2014privacy, fredrikson2015model}. Attribute inference attacks assume that an attacker possesses information about non-sensitive attributes of the data record and has access to the output of the trained model. Specifically, when a text instance has a missing attribute with $t$ possible values, the attacker constructs $t$ different input vectors and feeds them into the target model. The model's perplexity is then used to select the input most likely to be a member of the target model's training dataset~\cite{yeom2018privacy}. For example, ~\cite{lehman2021does} use a simple probing attack on BERT, trained on MIMIC-III patient notes, using a fill-in-the-blank template to recover patient names and their associated conditions. 

\myparagraph{How Attribute Inference Attacks on LLMs work?} ~\cite{chen2021killing} steal the functionality of a BERT-based API fine-tuned on TP-US, AG news and Blog datasets through only API access and then reveal sensitive attributes of the training data from the stolen model. For each attribute of interest (\eg gender, age), they train an attribute inference model consisting of a multi-layer feed-forward network and a binary classifier, which takes the shadow BERT embeddings of a sequence with missing attributes as the input and emits the predicted attribute. The measure is the ability of the model to predict the attributes correctly. They find this to be effective versus using embeddings from the BERT base. They test different architectures, such as RoBERTa and XLNET, as the victim models and find the attack is alleviated, showing the importance of matching architectures (and pretraining). They find XLNET-large is more vulnerable than the smaller XLNET-base, possibly as the larger model is more well-generalized. Their attack accuracy is over 80\% on all datasets when assuming access to queries from the original data distribution as the victim model. Still, it falls to as low as 50\% when queries are from a dataset with a different distribution.  On the other hand, ~\cite{staab2023beyond} proposes the threat model of using available data online authored by users, such as posts on online social forums. These posts are integrated into a prompt template, prompting an LLM to discern the personal characteristics of the individual who authored them. They stage this attack on GPT-4, Llama, PaLM, and Claude, using Reddit profiles they curate to encompass personal attributes such as age, education, gender, occupation, relationship status, location, birthplace, and income. They achieve an accuracy of 84.6\% across all attributes using GPT-4. This poses a significant privacy risk, showing LLMs infer collections of personal author attributes 
from varied sources across unstructured text. Notably, it is possible for humans to collate such information on an individual as they compare their results to human annotators. Generally, they found LLM results were in line with human difficulty ratings for acquiring specific information about someone. However, LLMs allow faster and more accurate collation of information. They show a trend in increased accuracy of AI attacks with increased model size, and they find this trend persists across model families.

\section{LLM Defence Systemization}

Existing defences (summarised in Table~\ref{tab:llm-defences} and Table~\ref{tab:llmdefences-eval}) can be categorised into three groups: i) Data Pre-processing (i.e., sanitizing data, see Section~\ref{sec:pre}); ii) Privacy-Preserving Training (i.e., differentially private training, fine-tuning, compression, federated training, and in-context learning algorithms, see Section~\ref{sec:privatising_model_training}); and iii) Forgetting approaches (discussed in Appendix~\ref{app:post}). Each of these can be done with formal guarantees of differential privacy or without provable privacy guarantees.

\begin{table*}
\centering
\caption{Systematization of LLM defences in terms of defence approach, phase under defence, and LLM architecture.}
\small
\label{tab:llm-defences}
\begin{tabular}{l|l|ccc|cccc|ccc}
\hline
\multirow{2}{*}{\textbf{Year}} & \multirow{2}{*}{\textbf{Ref}} & \multicolumn{3}{c|}{\textbf{Defence}} & \multicolumn{4}{c|}{\textbf{Phase}} & \multicolumn{3}{c}{\textbf{Architecture}}\\
 &  & \textbf{Pre} & \textbf{Train} & \textbf{Post} & \textbf{PT} & \textbf{FT} & \textbf{ICL} & \textbf{FL} &  \textbf{Enc.} & \textbf{Dec.} & \textbf{Enc.-Dec.} \\ \hline
2021 & \cite{basu2021benchmarking} & - & \checkmark & - & - & \checkmark & - & \checkmark & \checkmark & - & - \\ \hline
2021 & \cite{hoory2021learning} & - & \checkmark & - & \checkmark & - & - & - & \checkmark & - & - \\ \hline
2021 & \cite{anil2021large} & - & \checkmark & - & \checkmark & - & - & - & \checkmark & - & - \\ \hline
2021 & \cite{yu2021large} & - & \checkmark & - & - & \checkmark & - & - & \checkmark & - & -  \\ \hline
2022 & \cite{kandpal2022deduplicating} & \checkmark & - & - & \checkmark & - & - & - & - & \checkmark & - \\ \hline
2022 & \cite{lee2021deduplicating} & \checkmark & - & - & \checkmark & - & - & - & - & \checkmark & - \\ \hline
2022 & \cite{dupuy2022efficient} & - & \checkmark & - & - & \checkmark & - & - & \checkmark & - & - \\ \hline
2022 & \cite{li2021large} & - & \checkmark & - & \checkmark & - & - & - & \checkmark & \checkmark & - \\ \hline
2022 & \cite{yu2021differentially} & - & \checkmark & - & - & \checkmark & - & - & \checkmark & - & - \\ \hline
2022 & \cite{mireshghallah2022differentially} & - & \checkmark & - & - & \checkmark  & - & - & \checkmark & - & - \\ \hline
2022 & \cite{ponomareva2022training} & - & \checkmark & - & \checkmark & - & - & - & - & - & \checkmark \\ \hline
2022 & \cite{ginart2022submix}  & - & - & \checkmark & \checkmark & - & - & - & - & \checkmark & - \\ \hline
2022 & \cite{majmudar2022differentially} & - & - & \checkmark & \checkmark & - & - & - & \checkmark & - & - \\ \hline
2022 & \cite{jang2022knowledge} & - & - & \checkmark & \checkmark & - & - & - & - & \checkmark & - \\ \hline
2022 & \cite{carlini2022quantifying} & \checkmark & - & - & \checkmark & - & - & - & - & \checkmark & -  \\ \hline
2023 & \cite{xu2023training} & - & \checkmark & - & - & - & - & \checkmark & \checkmark & - & - \\ \hline
2023 & \cite{ishibashi2023knowledge}  & - & - & \checkmark & - & \checkmark & - & - & - & \checkmark & - \\ \hline
2023 & \cite{wu2023depn} & - & - & \checkmark & \checkmark & - & - & - & \checkmark & - & - \\ \hline
2023 & \cite{kassem2023preserving} & - & - & \checkmark & \checkmark & - & - & - & - & \checkmark & - \\ \hline
2024 & \cite{srivastava2024amplifying} & - & \checkmark & - & - & \checkmark & - & - & - & \checkmark & - \\ \hline
\end{tabular}
\end{table*}

\begin{table*}
\centering
\caption{Systematization of LLM defences in terms of the type of the model, task, training dataset and defence success metrics.}
\footnotesize
\label{tab:llmdefences-eval}
\begin{tabular}{l|l|p{20mm}|l|p{20mm}|p{30mm}} 
\hline

\textbf{Year}  &  \textbf{Ref} & \textbf{LLM Arch.} & \textbf{LLM Task}  &  \textbf{Data} & \textbf{Metrics} \\ 
\hline
2021 & \cite{basu2021benchmarking} & BERT, ALBERT, RoBERTa, DistilBERT & Text classification & Depression dataset$\ddag$ & Acc \\ \hline
2021 & \cite{hoory2021learning} & BERT & MLM & MIMIC-III, Wikipedia, BooksCorpus & Downstream Acc, Secret Exposure \\ \hline
2021 & \cite{anil2021large} & BERT-Large & MLM & Wikipedia, Books corpus & Acc \\ \hline
2021 & \cite{yu2021large} &  BERT & Multiple & GLUE (MNLI, QQP, QNLI, and SST-2) & Acc, MIA success rate \\ \hline
2022 & \cite{kandpal2022deduplicating}  & GPT-2 style & Next-token Pred & OpenWebText, C4 & \# Duplicates \\ \hline
2022 & \cite{lee2021deduplicating}  & Transformer XL & Next-token Pred & C4 & \# Duplicates \\ \hline
2022 & \cite{dupuy2022efficient} & BERT & Text classification & ATIS, SNIPS, NLU-EVAL & semantic error rate, MIA AUC \\ \hline
2022 & \cite{li2021large}  & BERT, RoBERTa, GPT-2 & Mixed & GLUE (MNLI, QQP, QNLI, and SST-2) & BLEU, ROUGE-L, PPL, F1 \\ \hline
2022 & \cite{yu2021differentially}  & RoBERTa-Large & Text classification & MNLI & BLEU, ROUGE-L,PPL \\ \hline
2022 & \cite{mireshghallah2022differentially} & BERT & Multiple & GLUE (MNLI, QQP, QNLI, SST-2) & Acc \\ \hline
2022 & \cite{ponomareva2022training} & T5 & Next-token Pred & C4 & Downstream Acc, Sequence Matching \\ \hline
2022 & \cite{ginart2022submix}  & GPT-2 & Next-token Pred & Wikitext-103, BigPatent-G & privacy loss \\ \hline
2022 & \cite{majmudar2022differentially} & RoBERTa-style & MLM & Common Crawl, Wikipedia, mC4 & PPL \\ \hline
2022 & \cite{jang2022knowledge}& GPT-Neo, OPT & Next-token Pred & The Pile & Memorization Acc, Empirical Definition of Forgetting \\ \hline
2022 & \cite{carlini2022quantifying}  & GPT-Neo, GPT-2 & Next-token Pred & The Pile & \# Duplicates \\ \hline
2023 & \cite{xu2023training}  & FOFE-like & Encoding & Stack Overflow & PPL \\ \hline
2023 & \cite{ishibashi2023knowledge}  & LLaMA, GPT-J & Mixed & LM Evaluation Harness & leakage rate, PPL \\ \hline
2023 & \cite{wu2023depn}  & BERTbase & MLM & Enron dataset & exposure, PPL \\ \hline
2023 & \cite{kassem2023preserving}  & GPT-Neo, OPT & Next-token Pred & The Pile & N-SacreBLEU, PPL \\ \hline
2024 & \cite{srivastava2024amplifying} & DistilGPT-2 & Next-token Pred & Wikitext-103 & Bias \\ \hline
\end{tabular}
\end{table*}

\section{Data Pre-processing Approaches}
\label{sec:pre}
\myparagraph{Sanitizing training data}. Sanitization aims to remove all sensitive information from data before model training~\cite{aura2006scanning, dernoncourt2017identification} and is widely applied in healthcare text to remove Personally Identifiable Information (PII). However, for large text datasets, identifying and removing specific sensitive text sequences is only possible through automated methods such as pattern-based parsers. \textbf{Formally defining private information in Natural Language is inherently complex~\cite{brown2020language}, making it challenging to design an automated sanitization method to guarantee the removal of all potentially sensitive sequences.} Data sanitization approaches are practical when sensitive information follows a context-independent, consistent format (e.g., national security numbers etc.), that is often unrealistic in real-world scenarios. The effectiveness of data sanitization in preserving privacy cannot be precisely measured or guaranteed, indicating that it should not be relied upon as the sole privacy-preserving measure for text data.

\myparagraph{Deduplicating training data.} There is often significant sequence-level duplication within and across text documents. It has been shown that removing duplicate sequences from the training data, using a suffix array-based algorithm, results in GPT-2 regenerating approximately 10x less training data~\cite{lee2021deduplicating}. Building on this, it has been empirically demonstrated that the likelihood of a GPT-Neo generating exact sequences from the training data scales with training data duplicates~\cite{carlini2022quantifying}. Removing exact duplicate sequences from the training data also protected GPT-2 from extraction attacks without reducing model performance~\cite{kandpal2022deduplicating}. A suite of decoder LLMs trained on sequence deduplicated training data are significantly less susceptible to MIAs~\cite{duan2024membership}. However, approximate memorization as measured by edit distance (SacreBLEU score~\cite{post2018call}) is still high in LLMs pretrained with sequence deduplicated training data~\cite{kassem2023preserving}. \textbf{Data deduplication effectively safeguards against training data leakage from LLMs but focuses on reducing memorization on
average and cannot guarantee to prevent memorization of a specific training example.}

\section{Privacy-Preserving Training}
\label{sec:privatising_model_training}

Private training approaches aim to train LLMs to preserve the privacy of their training data whilst also maintaining utility. 

\subsection{Differential Privacy} 

The framework of differential privacy (DP) provides formal privacy guarantees~\cite{dwork2014algorithmic}. An algorithm is DP if its outputs are statistically indistinguishable on neighbouring datasets that differ by only one record. Formally, a randomised algorithm $\mathcal{A}$ is ($\varepsilon, \delta$)-DP if for any neighbouring datasets, $D, D'$ that differ in exactly one record and for all $S \in Range(\mathcal{A})$, $Pr[\mathcal{A}(D)\in S] \leq e^{\varepsilon}Pr[\mathcal{A}(D')\in S]$, where $\varepsilon$ is the privacy budget that sets an upper bound on the potential privacy leakage in the worst-case scenario. A smaller $\varepsilon$ indicates stronger privacy protection. The parameter $\delta$ is in the order of the inverse of the dataset's size, so it is typically very small.

\myparagraph{DP Stochastic Gradient Descent (DP-SGD)} is the de facto strategy for training ML models under the framework of DP~\cite{abadi2016deep, song2013stochastic, bassily2014private}. In each training iteration, DP-SGD introduces two modifications to vanilla SGD: (1) the gradients for individual examples are clipped to a fixed norm, $C$, to limit the influence of individual training examples on model updates, and (2) calibrated Gaussian noise, proportional to $C$, is added to the aggregated clipped gradients. This noisy gradient is then utilized to update the model parameters. The amount of noise added, relative to the clipping norm, determines the strictness of the upper limit $\varepsilon$ on privacy loss that can be ensured. 

\subsection{Federated Learning} 
Federated Learning (FL)~\cite{mcmahan2017communication} is a technique to avoid directly exchanging private data. A model is trained through local updates from individual clients (\eg devices or organisations), where the client only sends model weights (never raw data) which are combined at a central server. FL is pivotal in several LLM applications, such as training LLMs on electronic health records from different hospitals. Early work in smaller LMs (LSTMs) demonstrated that memorization in an FL setting is reduced by 60\% or more compared to central training~\cite {thakkar2020understanding}. However, model updates during FL, including gradients and parameters, can be deliberately exploited to unveil sensitive information~\cite{zhu2019deep, deng2021tag, gupta2022recovering, fowl2022decepticons, rashid2023fltrojan, li2023beyond}. 

\myparagraph{Combining DP with FL.} Combining FL with DP (DPFL) provides guaranteed privacy protection and was first introduced in DP-Federated Averaging (DP-FedAvg)~\cite{mcmahan2017communication}. This involves using local SGD on each client alongside a server performing model averaging. Later works extended the above work by applying DPFL to LSTMs on millions of real devices for mobile keyboard prediction~\cite{hard2018federated, ramaswamy2020training}. However, these LMs have an order of magnitude fewer parameters and a smaller vocabulary size compared with today's LLMs. Notably, challenges of applying DPFL to LLMs include (1) the high number of parameters of LLMs results in a massive payload, putting communication strain on the server; (2) the large vocabulary size means the distribution of word frequency in natural language exhibits a pronounced skewness and long-tailed distribution, so the signal-to-DP noise ratio for rare words is very small, causing a loss of information during DP training. 

\subsection{Analysis of Privacy-Preserving Training}

\myparagraph{How does LLM size affect Privacy-Preserving Training?} Computing per-example gradients in DP-SGD incurs substantial memory and computational overhead, posing significant challenges for LLMs with their large parameter count and vocabulary size. A research focus in LLMs has hence been improving the efficiency of DP-training whilst maintaining accuracy~\cite{anil2021large,hoory2021learning,ponomareva2022training, yu2021differentially, yu2021large,li2021large, dupuy2022efficient}. 

A number of studies investigate the increasing efficiency of DP-SGD during pre-training LLMs~\cite{anil2021large,hoory2021learning,ponomareva2022training}. They found scaling batch sizes during training to millions (mega-batches)~\cite{anil2021large,hoory2021learning} and using an increasing batch-size schedule~\cite{anil2021large} achieved higher accuracy and faster pre-training under DP-SGD. However, accuracy was up to 10\% lower compared with non-private standards~\cite{hoory2021learning,anil2021large}. JAX and XLA have also been employed for faster training under DP~\cite{ponomareva2022training}. 

DP private tokenization has also been explored as part of the DP-training pipeline in LLMs~\cite{hoory2021learning, ponomareva2022training}. This involves injecting noise into the word histogram of the corpus. Interestingly, some of the drop in utility from DP-training can be recovered by using the DP tokenizer instead of a non-DP tokenizer~\cite{ponomareva2022training}.

Another set of studies investigates efficient DP-SGD for fine-tuning LLMs~\cite{yu2021large,  yu2021differentially, li2021large, dupuy2022efficient}. A number of approaches explore variants of parameter-efficient approaches~\cite{yu2021large, yu2021differentially} such as Low-Rank Adaptation (LoRA)~\cite{hu2021lora}. These serve to reduce the number of tuneable parameters in the model, enhancing efficiency. These approaches have been applied to BERT~\cite{yu2021large} and RoBERTa-Large~\cite{yu2021differentially} fine-tuned on GLUE tasks and achieved within 5\% accuracy of non-private baselines at strong privacy guarantees of $\varepsilon \leq 8$. Parameter-efficient methods have also been employed for DPFL~\cite{xu2023training}. This is combined with partial embedding updates, where each client only transmits scaled updates to the central server for multiple randomly selected words. This helps to reduce the noise introduced by DP across sparse embeddings. 

A parallel line of work on BERT, RoBERTa and GPT-2 finds larger batch sizes also allow for efficient DPSGD, akin to DP pretraining. Combining this with a careful selection of hyperparameters, including higher learning rate and smaller clipping limits, can even outperform non-privately fine-tuned baselines on downstream tasks with stringent privacy guarantees ($\varepsilon \in {3, 8}$). Clipping the gradient on micro-batches can adapt DP training of LLMs to GPU and improve DPSGD training speed~\cite{dupuy2022efficient}. Interestingly, smaller fine-tuned LM performance is degraded faster across DP-training than larger LLMs~\cite{basu2021benchmarking}. This enabled training on resource-constrained devices.  

\myparagraph{In which phase of building LLMs we can apply DP?} DP has been investigated for pretraining~\cite{anil2021large,hoory2021learning,ponomareva2022training} and fine-tuning~\cite{yu2021large, yu2021differentially, li2021large, dupuy2022efficient, basu2021benchmarking} LLMs. DP-pretraining does not significantly impact LLM downstream task performance~\cite{hoory2021learning, ponomareva2022training}. However, a thorough study of the interaction of DP pretraining and fine-tuning has not been conducted. DP has also been investigated for compressing LLMs by using DP-SGD to train a teacher model and then performing knowledge distillation, resulting in DP student models ~\cite{mireshghallah2022differentially}. These come within 0.8\% accuracy of KD with a non-DP fine-tuned teacher under a privacy budget of $\varepsilon=4.25$. 

\looseness=-1 \myparagraph{How does LLM data affect Privacy-Preserving Training?} In a benchmarking study, it was found that models training on fine-tuning training corpora have more performance degradation when trained with DP~\cite{basu2021benchmarking}. In the real world, fine-tuning datasets are often small and more likely to contain private information. In these cases, the trade-off with utility could be problematic. Additionally, a concern is that noise is added equally to all data points during DPSGD, which means for rare groups in the data, convergence will be more difficult. If training on tasks with imbalanced data DP will reduce minority group performance more. For example, DP fine-tuning of DistilGPT-2 amplifies bias concerning race, gender and religion~\cite{srivastava2024amplifying}. Techniques to address bias (\,  e.g. counterfactual data augmentation) could be combined with DP to help mitigate this effect. 

\section{Post-training Approaches}


\subsection{Privacy-Preserving Inference}

Privacy-preserving inference aims to protect private information in the training data from leakage in model outputs at inference time. DP can be applied to inference in that, on an adversarially selected feature, a model's prediction does not vary too much whether or not a certain single data point is in the training set.
A, is ($\epsilon$)-differentially private if for any two datasets $D, D'\in (X × Y)^n$ differing in exactly one entry, for all $x \in X$, and all sets $Y \subseteq \gamma $: 
\begin{equation}
    Pr[f(x; A(D))] \leq e^{\epsilon}Pr[f(x; A(D')) \in Y ]
\end{equation}

One approach uses an ensemble of LLMs fine-tuned on disjoint part of the private data~\cite{ginart2022submix}. During inference, if all models agree on the next-token distribution, this intuitively means it is not possible for the next token to leak sensitive information from the training data. However, if high disagreement is observed between models, the predictions are mixed with those of the public pre-trained model (that the fine-tuned models were initialised from) to minimize privacy leakage. On GPT-2 fine-tuned on Wikitext-103 and BigPatent-G datasets, privacy leakage was as small as $\epsilon=2$ but utility (perplexity) of almost 75\% of non-private fine-tuning was maintained. However, decoding from an ensemble of LLMs can produce incoherent text. In addition, this approach increased storage and computation 8-fold. A possible solution is to fine-tune only the last layer of the ensemble models so they share more layers. 

A more lightweight approach is a perturbation mechanism for the trained model at the decoding stage, which can be used with any LLM~\cite{majmudar2022differentially}. A perturbed output distribution is obtained by linear interpolation between the original model output distribution for a given input and the uniform distribution. The perturbed probability can be used to randomly select a token from the vocabulary to fill the mask. However, on RoBERTa-style trained on MLM on Common Crawl, Wikipedia, and mC4, model performance significantly degraded at privacy guarantees below $\epsilon = 60$. 

\subsection{Machine Unlearning}

Regulations implemented by the General Data Protection Regulation (GDPR) in the EU or the California Consumer Privacy Act (CCPA) in the US have incorporated clauses regarding the right to be forgotten. These provisions mandate that industry applications delete data associated with an individual from their systems. This also applies to machine learning model applications. Complying with such mandates post-training by continuously removing data and retraining does not scale. Exact unlearning involves retraining the model from scratch on a new dataset that does not contain the removal samples. However, approximate unlearning comprises a set of techniques to remove the influence of specific training examples from the weights of a trained model by producing a new set of weights that approximate the weights from retraining. This is particularly significant for LLMs, where retraining is highly costly and impractical. There is a growing body of work on unlearning in machine learning models, with a Challenge on this topic for Computer Vision revealed in 2023\footnote{\url{https://unlearning-challenge.github.io/}}. Unlearning in LLMs has been explored by a number of studies~\cite{jang2022knowledge, kassem2023preserving, eldan2023s, chen2023unlearn, pawelczyk2023context}. 

One approach is to minimise the negative log-likelihood, reversing the training objective during language modelling to forget specific examples (the ``forget set") from the original training data~\cite{jang2022knowledge}. For GPT-Neo 2.7B, this method effectively protects the ``forget set" from extraction attacks, with minimal degradation to performance. The success of forgetting was measured by measuring the memorization accuracy and susceptibility to extraction attacks of a point. A point is considered forgotten if it has a lower forgetting threshold, calculated as the average memorization score of 10,000 points not seen during training. A limitation of this forgetting threshold is that it depends on which data samples are chosen. The approach degraded the fluency and coherency of generated suffixes, particularly for the larger GPT-Neo 125M, which severely degraded performance on downstream classification and dialogue tasks. Notably, OPT LMs (duplicated during training) have lower memorization than GPT-Neo LMs, confirming that deduplication of training corpora helps mitigate privacy risks. Larger LMs require fewer epochs to forget sequences, suggesting larger LMs are stronger ``unlearners". Also, unlearning at larger sample sizes (up to 128) significantly degraded model performance. 

Another approach uses a reinforcement learning feedback loop to forget seuqences~\cite{kassem2023preserving}. Given the prefix and suffix of the training instance to forget from the model, the prefix is input to the LLM and generates a suffix. The negative BERTScore~\cite{zhang2019bertscore} is computed to measure the dissimilarity between the true suffix and the generated suffix. The LLM is then fine-tuned using the dissimilarity score as a reward signal to incentivise the unlearning of the training point. Performance degradation was minimal on common text classification benchmarks, and memorization was significantly decreased compared with reversing the training objective~\cite{jang2022knowledge}. Notably, larger models tend to forget memorized data faster. Combining this approach with data deduplication enhanced privacy further, demonstrating the effectiveness of combining privacy-preserving approaches. Higher perplexity (lower utility) was observed in certain examples, likely attributed to optimizing for the dissimilarity of sequences only and not also perplexity. Multi-objective reinforcement learning could provide a route to enhance this approach. 

\subsection{Model Editing}

Model Editing approaches aim to update a model’s behaviour concerning a specific edit descriptor by updating, erasing or inserting knowledge. There have been numerous works on machine editing in LLMs~\cite{de2021editing, meng2022locating, huang2023transformer, dai2021knowledge}. Based on the idea that the feed-forward network module within the Transformer architecture can be likened to a key-value memory, where each key signifies a text pattern and each value denotes a distribution across the vocabulary~\cite{geva2020transformer}, locate-then-edit techniques propose identifying which parameters in the network store specific knowledge and modifying them to edit these~\cite{geva2020transformer,dai2021knowledge, meng2022locating, meng2022mass, chang2023localization}. Private information from the training data could be stored in specific neurons. This means they could remove this information memorized by the model by detecting and deleting these privacy neurons in a locate-then-edit approach. 

To estimate the privacy attribution scores for private information, gradient integration can be used to compute the contributions of multiple markers to neuron activations simultaneously~\cite{wu2023depn}. The top z neurons with the highest attribution scores can have their activations set to zero to erase the model memorization of the corresponding private information. Overall, the goal is to reduce the probability of privacy leakage of a sequence containing private information, given the context. On BERTbase, this method reduced privacy leakage risk compared to BERTbase with fine-tuning by almost 40\%~\cite {wu2023depn}. However, the leakage risk is still higher than using DP-fine-tuning. However, they note this approach is significantly less computationally expensive than DP training. Their approach maintains model performance close to the original performance on the MLM task, compared to DP training. However, more neurons need to be edited for greater data erasure, and this causes significant drops in model performance and also reduces the effectiveness of reducing privacy risk. 

Another approach is to minimally fine-tune already trained LLMs to generate safe responses when queried about a particular knowledge domain~\cite{ishibashi2023knowledge}. This can be done by fine-tuning the model on question-answer pairs and replacing the sensitive answer with a safe response in the data. Fine-tuning just the weight matrices in the MLP layers using LoRA is sufficient. On LLaMA and GPT-J, the model's performance on the target removal data reduces from 74\% to 0\% accuracy after private fine-tuning and around 80\% of the outputs generated in response to training queries were changed into sanitization phrases. A qualitative assessment using extraction attacks shows the method reduces extraction attack risk. However, if applied widely, it could reduce utility in practice due to the stringent filtering.

\section{Discussion \& Future Opportunities}

\myparagraph{LLM Size \& Emergent Behaviours.} The majority of studies on LLM attacks and defences focus on GPT-2-like and BERT-like models, largely due availability and practicalities of training. However, these are an order of magnitude smaller than many LLMs today. It is not clear if current privacy trends will extrapolate to much larger LLMs. Future research is needed to analyse emergent behaviour in larger LLMs and the effect on privacy attacks and defences. Additionally, the studies analysed here have focused primarily on the encoder and decoder transformer-based LLMs, finding the latter to be the most vulnerable to attacks in general. Emerging more efficient architecture variations (e.g., Mamba~\cite{gu2023mamba}) are an interesting avenue to explore in terms of effects on memorization and training data leakage and finding architectures that inherently reduce memorization. 

\myparagraph{Privacy Risks of ICL.} Using LLMs available via APIs as few shot learners is becoming increasingly common. This introduces new risks around data privacy as LLMs may leak or regurgitate sensitive data given as context in prompts~\cite{priyanshu2023chatbots, duan2023privacy, tang2023privacy}. Concerningly, it has also been demonstrated that even very capable generative LLMs do not well understand private context~\cite{mireshghallah2023can}. Prompt data used for prompt-tuning in LLMs is highly susceptible to privacy attacks~\cite{duan2023privacy, zhang2023ethicist}. An interesting emerging research avenue is Privacy-preserving ICL, with initial works focused on DP-synthetic prompts~\cite{tang2023privacy} and DP-outputs by consensus from an ensemble of LLM responses~\cite{panda2023differentially}. 

\myparagraph{New avenues for Defence.} Multi-step reasoning encompasses a set of approaches which facilitate more effective reasoning by guiding LLMs to produce a sequence of intermediate steps before giving the final answer~\cite{wei2022chain, yao2023tree}. Investigating reasoning approaches as privacy-preserving methods at inference time~\cite{mireshghallah2023can} offers an exciting and open avenue for research. Another interesting approach is ICL for machine unlearning, such as constructing prompt contexts which induce model behaviour indistinguishable from a model re-trained without certain datapoints~\cite{pawelczyk2023context}. Further work is needed to explore effective privacy-preserving prompt contexts to ensure the safety of context data. 

\myparagraph{Privacy Risks of Multimodal Approaches.} A cause for concern is Multimodal LLMs~\cite{yin2023survey}, which hold the risk of leaking private information across data modalities. Especially as they are often applied in situations with highly sensitive data, such as medical report generation or disease identification. Multimodal models have been demonstrated to be vulnerable to MIAs in an image-to-text model~\cite{hu2022m}. This remains a largely unexplored area and is of particular importance regarding patient data. 

\myparagraph{Emerging Risks to wider LLM Systems.} There are several emerging themes from Trustworthy AI which are concerning for LLMs. Privacy Backdoors,  tampering with a model's weights to allow an attacker to reconstruct data during future training, allow for stronger privacy attacks even on models trained under DP~\cite{feng2024privacy}. It is common practice to download pre-trained LLMs from public repositories and fine-tune them, presenting a significant risk of backdoor attacks. Backdoors can also be introduced into models available via prompt-tuning APIs, for example, by incorporating backdoors into the prompt context~\cite{zhao2024universal, zhao2023prompt}. Further understanding the risks of privacy backdoors to data privacy across LLM development phases is an important research avenue. Privacy Side Channels refer to system-level components of models which can be exploited to improve privacy attack success rates, including attacks targeting data preprocessing approaches, tokenizers, output filtering and query filtering~\cite{debenedetti2023privacy}. For example, knowing data will be deduplicated, an attacker can infer a data point in the training data by detecting whether near duplicates were deleted. Inserting poison examples into the dataset before deduplication amplifies these dependencies and increases MIA strength. Likewise, output filters, used to prevent verbatim training data from being regurgitated from a model, can actually facilitate near-perfect MIAs~\cite{debenedetti2023privacy}. Many such components commonly employed by modern LLM systems to improve privacy may actually increase privacy risks and so are a serious area for concern. Further investigation into these privacy risks and associated mitigations in LLMs is essential to provide up-to-date guidance to practitioners. 

\myparagraph{Conclusion} We present a privacy attack and defence framework for data privacy in LLMs. We outline the current state of research and opportunities for privacy practitioners. Our frameworks and taxonomies aim to provide a foundation with which to understand and categorise data privacy research on LLMs. 

\section*{Acknowledgments}
For helpful advice and feedback, we would like to thank Niloofar Mireshghallah and Youmna Hashem.

AW acknowledges support from a Turing AI Fellowship under grant EP/V025279/1, and the Leverhulme Trust via CFI.

VS acknowledges support from a UCL UKRI Centre for Doctoral Training in AI-enabled Healthcare studentship (EP/S021612/1), a NIHR Biomedical Research Centre at University College London Hospital NHS Trust studentship, and an Alan Turing Institute Community Award. 

\bibliographystyle{IEEEtran}
\bibliography{main}  

\end{document}